\documentclass[letterpaper, 10 pt, conference]{ieeeconf}  

\IEEEoverridecommandlockouts                              

\usepackage{graphicx} 
\usepackage{amsmath} 
\usepackage{bbm}
\usepackage[%
  ruled                             
]{algorithm2e}    
\usepackage{subfig}
\usepackage[table,xcdraw]{xcolor}
\usepackage{wrapfig}
\usepackage{hyperref}
\usepackage{cleveref}
\usepackage{multirow}

\newcommand\abs[1]{\left|#1\right|}

\title{\LARGE \bf
Building a Library of Tactile Skills Based on FingerVision
}

\author{Boris Belousov*, Alymbek Sadybakasov*, Bastian Wibranek, Filipe Veiga, Oliver Tessmann, and Jan Peters%
\thanks{*Authors contributed equally}%
\thanks{B.B., A.S., and J.P. are with Intelligent Autonomous Systems Lab,
	TU Darmstadt, Germany {\tt\small surname@ias.tu-darmstadt.de}}%
\thanks{J.P is in addition with MPI for Intelligent Systems, Tübingen}%
\thanks{B.W. and O.T. are with Digital Design Unit,
	Technische Universität Darmstadt, Germany
	{\tt\small surname@dg.tu-darmstadt.de}}%
\thanks{F.V. is with CSAIL, MIT, USA {\tt\small fveiga@csail.mit.edu}}%
\thanks{This project has received funding
from the European Union’s Horizon 2020 research and innovation programme under grant agreement No 640554.}%
}

\begin{document}
\maketitle
\thispagestyle{empty}
\pagestyle{empty}

\begin{abstract}
Camera-based tactile sensors are emerging as a promising inexpensive solution
for tactile-enhanced manipulation tasks.
A recently introduced FingerVision sensor was shown capable of generating reliable signals for
force estimation, object pose estimation, and slip detection.
In this paper, we build upon the FingerVision design, improving already existing control algorithms,
and, more importantly, expanding its range of applicability to more challenging tasks
by utilizing raw skin deformation data for control.
In contrast to previous approaches that rely on the average deformation of the whole sensor surface,
we directly employ local deviations of each spherical marker immersed in the silicone body of the sensor
for feedback control and as input to learning tasks.
We show that with such input, substances of varying texture and viscosity can be distinguished
on the basis of tactile sensations evoked while stirring them.
As another application, we learn a mapping between skin deformation and force applied to an object.
To demonstrate the full range of capabilities of the proposed controllers,
we deploy them in a challenging architectural assembly task
that involves inserting a load-bearing element underneath a bendable plate at the point of maximum load.
\end{abstract}

\section{INTRODUCTION}
Endowing robots with a sense of touch is a long-standing problem in robotics~\cite{cutkosky2008force}.
A variety of sensor designs were proposed in the past~\cite{dahiya2013directions}.
An early \emph{camera-based} solution~\cite{zielinska1996shear} measured shear force
by tracking reflective cones protruding from an opaque surface of a polymer.
To get a full 3D vector of force, the GelForce sensor~\cite{kamiyama2005vision}
used a camera to track two layers of colored dots immersed in a piece of elastomeric gel with an opaque coating.
The GelSight sensor~\cite{yuan2015measurement} has a similar design to GelForce
but in addition it can estimate 3D topography of a contact surface and it has a much higher resolution.
Unlike the previous sensors, which had an opaque cover and illuminated the sensor skin from within,
the FingerVision sensor~\cite{yamaguchi2016combining} is transparent and relies on external lighting for illumination.
Such design allows capturing object properties such as shape and color even before contact.
Moreover, slip detection turns into a well-studied computer vision problem of motion estimation.

In order to efficiently use a tactile sensor for robot control,
sensor data needs to be passed to a control system in an appropriate form.
A preprocessing pipeline for FingerVision together with a number of tactile feedback controllers
was proposed in~\cite{yamaguchi2017implementing}.
Various behaviors, such as \emph{gentle grasp}, \emph{handover}, \emph{in-hand manipulation},
and \emph{object tracking}, were demonstrated.
However, in all cases, force-based controllers relied on averaged force values
over all markers~\cite{yamaguchi2018fingervision}:
either the average force itself (object tracking),
or the average of the absolute force values (handover),
or some score based on the absolute force values (gentle grasp).

\begin{figure}[b]
\vspace{-1em}
      \centering
      \includegraphics[scale=0.055,trim={0 0 0 2em},clip]{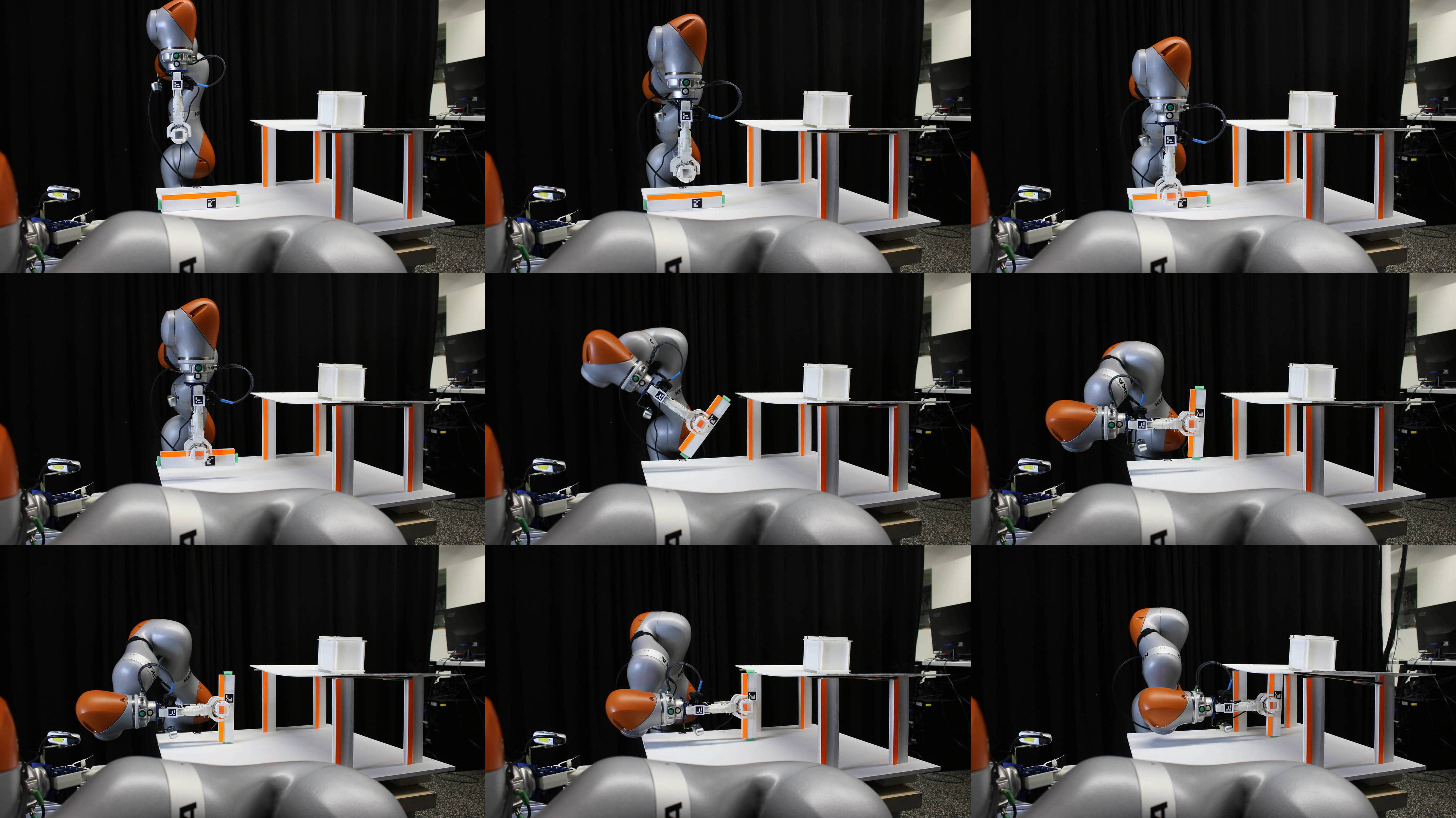}
      \caption{
      The architectural assembly task is comprised of a sequence of manipulations.
	After being located and scanned with the FingerVision camera,
	a load-bearing element is grasped and rotated.
	The robot then goes down until a contact with the ground surface is detected.
	Lastly, the point of the highest load is identified via continuous force measurement by FingerVision
	and the part is placed there.
      }
      \label{fig:AssemblyTask}
\end{figure}

In this paper, we extend the range of capabilities of the FingerVision sensor
by developing a library of controllers based on marker deviations and proximity vision,
which subsumes already existing tactile skills and adds a number of novel ones.
Our contributions are summarized in Table~\ref{table:tactile_skills}.
We show that feedback controllers utilizing local force information
can be designed to enable in-hand object rotation and whole-arm manipulation.
Furthermore, we demonstrate the usefulness of rich tactile feedback in two learning scenarios:
learning to apply a specific force to an object
and inferring texture and viscosity properties of a substance by stirring it.
Along with developing these novel skills, we also make a number of improvements
to already existing capabilities.
Namely, we improve the marker tracking algorithm by integrating a Kalman filter into the estimation procedure,
we extend the handover skill with a leaky integrator to allow for seamless gripper opening and closure,
we enhance the force tracking skill with a speed control feature to enable smoother hand-guiding behavior,
and we add a visual scan skill that utilizes proximity vision to locate an object boundary in case of objects that
do not fit inside the gripper.
Finally, we showcase the capabilities of the developed controllers
in an integrated architectural assembly task shown in Fig.~\ref{fig:AssemblyTask}.

\section{BUILDING CUSTOM FINGERVISION SENSOR}
\label{sec:building}
\begin{figure}[b]
\vspace{-1em}
\centering
  \subfloat{%
   \includegraphics[width=0.16\linewidth]{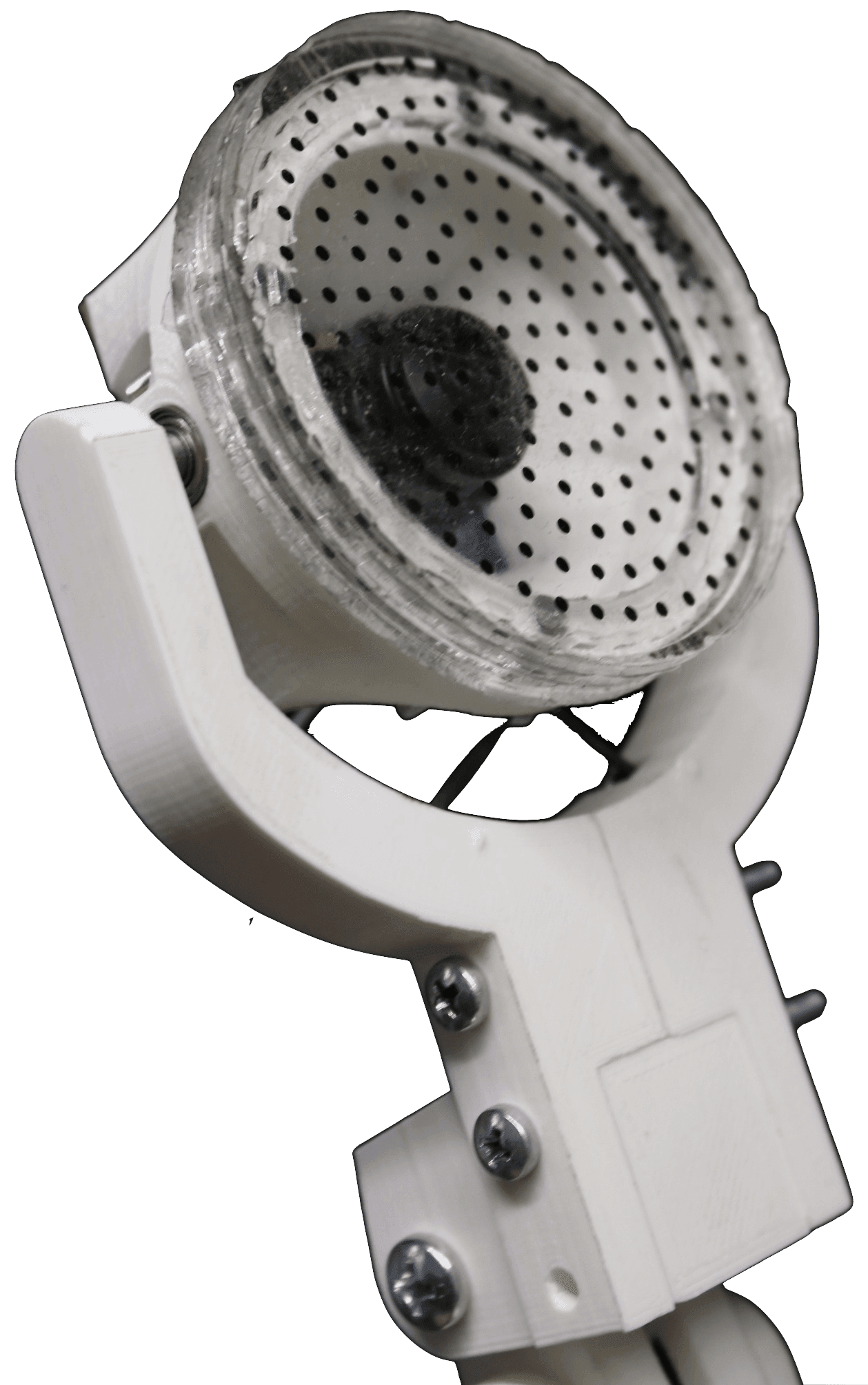}}
   \hspace{0.1em}
  \subfloat{%
   \includegraphics[width=0.19\linewidth]{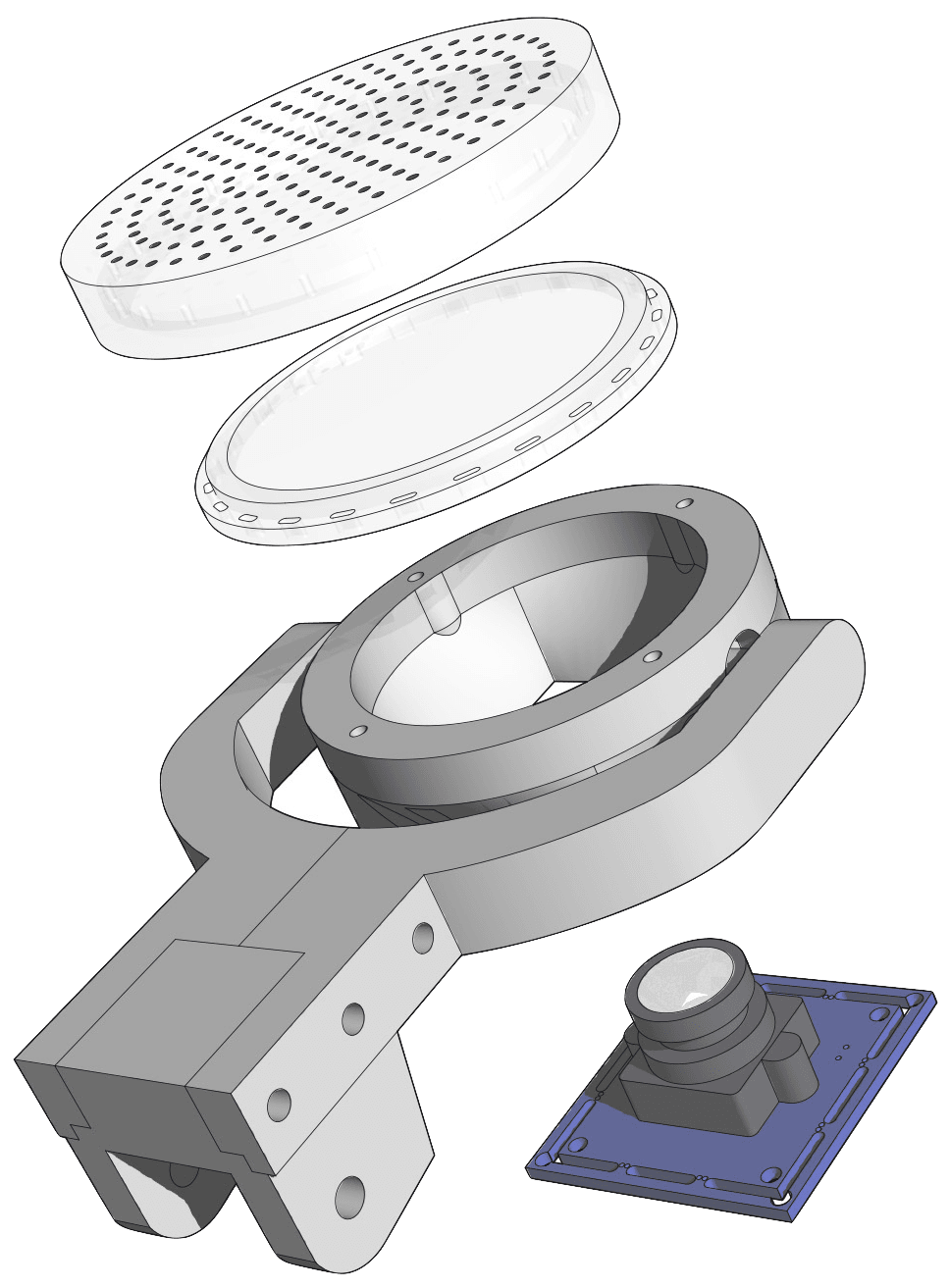}}
   \hspace{0.05em}
   \subfloat{%
   \includegraphics[width=0.3\linewidth]{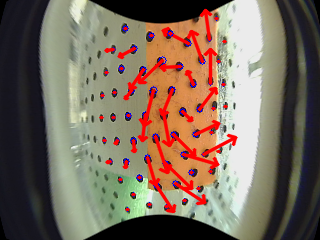}}
    \hspace{0.05em}
    \subfloat{%
   \includegraphics[width=0.3\linewidth]{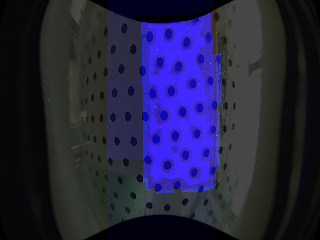}}
  \caption{Our FingerVision sensor has a round shape
  	allowing for more markers in the active area.
	Two major data modalities are shown:
	marker displacements (force estimation) and proximity vision (object pose and movement estimation).}
	\label{fig:finger_vision_sensor}
\end{figure}

The original FingerVision design, manufacturing instructions, and the data processing code
were presented in~\cite{yamaguchi2016combining}.
We make several modifications to accommodate for a different gripper
and available materials (see Fig.~\ref{fig:finger_vision_sensor}).
In particular, we use different silicone, markers, marker pattern, and housing.

Silicone Replisil 19 N glasklar was selected among alternatives for its transparency. However, it was too hard.
To increase the softness of the silicone, tactile mutator Slacker was added, as suggested in~\cite{gelsightmaster}.
Furthermore, to extend the time available for infusing the markers into the silicone,
additive Slo-Jo was employed, which granted up to 30 minutes dripping time.
Resulting material showed good durability: after several months of intensive experiments,
the silicone was still transparent and in good condition.
Direct comparison to the silicone used in the original FingerVision sensor was not carried out
due to its unavailability at the time of our experiments; however, we were able to reproduce
the results reported on FingerVision with our materials.

Using microbeads as markers, as envisaged by the original design~\cite{yamaguchi2016combining},
turned out to be problematic: the microbeads did not stick well and were shifting inside the silicone.
Bad adhesion might be due to the difference in silicone used.
As an alternative solution, iron oxide was mixed with the silicone to color it black,
and then the mix was injected through an acrylic template
to form spherical blobs inside the transparent silicone layer.
Such approach allowed for precise control of the positioning and shape of the injected markers.

Thanks to the fisheye lens of the FingerVision camera,
the density of tracked markers could be increased by giving the sensor body a round shape.
Fig.~\ref{fig:finger_vision_sensor} shows a schematic design that we used and its implementation.
Note the precision of marker placement and the radial symmetry of the pattern.

In our experiments, an under-actuated parallel gripper was employed.
Due to its specific morphology, an additional elastic degree of freedom between the sensor body
and the gripper phalange was added to enable grasping of objects
of different sizes (see Fig.~\ref{fig:finger_vision_sensor}).
Our manufacturing process is described in more detail in~\cite{alymbek_thesis}.

Two types of information were extracted from the FingerVision camera~\cite{yamaguchi2016combining}:
\emph{marker displacements} and \emph{proximity vision}.
Marker displacements, shown by red arrows in Fig.~\ref{fig:finger_vision_sensor},
provide local force estimates at marker locations.
Proximity vision delivers object pose and motion estimation information
via histogram-based background subtraction~\cite{yamaguchi2017implementing}.

\section{LIBRARY OF TACTILE SKILLS}
\label{sec:library}
\vspace{-.25em}
\begin{table}[t]
\vspace{.5em}
\caption{Tactile skills based on FingerVision. Most skills require 2 modalities,
	therefore they are arranged in a table with 4 basic modalities in rows and columns.
	Our new skills are shown in bold; italic highlights learning-based skills.}
\label{table:tactile_skills}
\vspace{-1em}
\begin{center}
    \begin{tabular}{|c|c|c|c|c|}
    \hline
    &  Markers & Force & \multirow{2}{*}{Object} & \multirow{2}{*}{Slip} \\
    & +\textbf{KF} \ref{subsec:kalman_blob} & +\textbf{KF} \ref{subsec:kalman_force}  &&\\
    \hline
	Markers & \textbf{ArmRot}   &  &  &  \\
   	 +\textbf{KF} & \ref{subsec:trq-arm-rot} &&& \\
    \hline
    Force& \textbf{\textit{ForceLearn}} & \multirow{2}{*}{GentleGrasp} &  &  \\
    	+\textbf{KF} & \ref{subsec:force-learn} &&& \\
    \hline
    \multirow{2}{*}{Object} & \textbf{InHandRot} & ForceTrack+ & ObjectTrack, &\\
    & \ref{subsec:in-hand-rot} & \textbf{SpeedCtrl} \ref{subsec:force-based-ctrl}
    	& \textbf{VisScan} \ref{sec:assembly} &  \\
    \hline
    \multirow{2}{*}{Slip} & \textbf{\textit{StirLearn}} & Handover+
    	& \multirow{2}{*}{InHandMan} & \multirow{2}{*}{Hold} \\
    & \ref{subsec:stir-learn} & \textbf{LeakyInt} \ref{subsec:leaky} && \\
    \hline
    \end{tabular}
\end{center}
\vspace{-2.5em}
\end{table}
We make a number of improvements upon the baseline skills introduced in prior
works~\cite{yamaguchi2016combining,yamaguchi2017implementing,yamaguchi2018fingervision}.
Novel skills and contributions are denoted in bold in Table~\ref{table:tactile_skills}.
Columns and rows correspond to sensor data modalities:
(i)~\emph{Markers} stands for marker displacements estimated using OpenCV~\cite{opencv_library},
(ii)~\emph{Force} stands for the averaged force computed based on \emph{Markers},
(iii)~\emph{Object} stands for object information such as shape, distance,
and orientation extracted from proximity vision,
(iv)~\emph{Slip} stands for optical-flow-based slip estimation.

Most skills require two modalities, therefore we arrange them in a 2D table.
Whenever there is a plus sign, it means an improvement to an existing skill.
Here is a brief summary of the improvements.
First, we add a \emph{Kalman filter} to the blob tracking algorithm,
which appreciably improves both individual marker tracking performance (Markers+\textbf{KF})
and force estimation (Force+\textbf{KF}).
Second, we enhance the \emph{Handover} skill with a \emph{leaky integrator}~\cite{veiga2018hand}
to enable seamless gripper opening and closure during object handover (Handover+\textbf{LeakyInt}).
Third, we extend the force tracking controller \emph{ForceTrack}
with a \emph{speed control} feature that enables smoother
and more natural force-based interactions with the robot (ForceTrack+\textbf{SpeedCtrl}).

Along with improvements to existing skills, we also introduce new capabilities.
First, building upon the \emph{Object} modality, we develop a \emph{visual scanning} skill \textbf{VisScan}
that can follow the shape of an object and find its boundary and size.
We demonstrate the utility of the VisScan skill in the assembly task (Sec.~\ref{sec:assembly}).
Second, we introduce a suite of skills based on raw marker displacement data:
(i)~the \emph{arm rotation} controller \textbf{ArmRot} makes use of circular vector field estimation,
which we demonstrate on rotation of an in-finger-held asymmetric heavy object;
(ii)~the \emph{in-hand rotation} skill \textbf{InHandRot} combines Markers and Object modalities
to accomplish in-finger object rotation, which illustrates the capacity of joint force and torque estimation;
(iii)~we demonstrate that end-to-end control based on marker displacement data is feasible
by developing two learning-based controllers---\textit{\textbf{ForceLearn}},
capable of associating tactile sensation (Markers) with applied force (Force),
which we showcase on pressing with a desired force in Newtons,
and \textit{\textbf{StirLearn}}, capable of associating tactile sensation (Markers)
with vision-based slip detection (Slip), which we showcase on identifying a substance through stirring.

\section{IMPROVED AND NOVEL TACTILE SKILLS}
\label{sec:improved-and-novel}
This section focuses on improved and newly introduced tactile skills.
Implementation details, evaluations, and example use cases are presented.
Subsequent sections cover learning-based skills (Sec.~\ref{sec:learning-based-skills})
and the showcase application in collaborative architectural assembly (Sec.~\ref{sec:assembly}).

\subsection{Blob Tracking Improved with the \textbf{Kalman Filter}}
\label{subsec:kalman_blob}
\begin{figure}[b]
\vspace{-1em}
\centering
    \includegraphics[width=0.95\linewidth,trim={0 1.5em 0em 1.5em},clip]{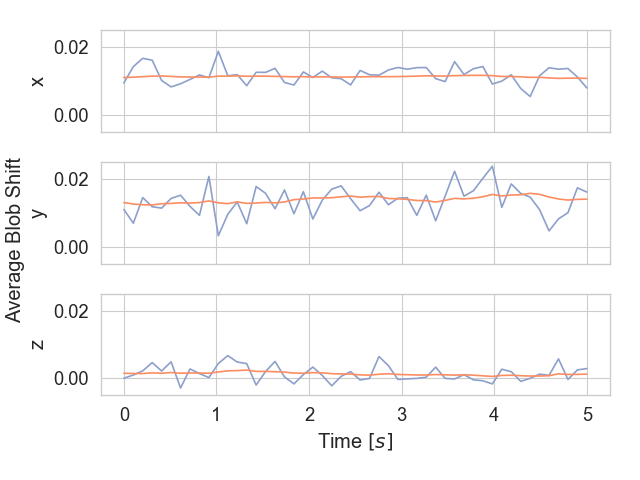}
    \caption{
    Blob tracking \emph{with} (orange) and \emph{without} (blue) the \textbf{Kalman filter} on a static object holding task.}
    \label{KalmanFilter}
\end{figure}
We start by describing the Kalman-filter-based enhancement
to the blob tracking algorithm of FingerVision~\cite{yamaguchi2016combining}.
Every marker in the silicone layer of the sensor is detected using OpenCV,
and its displacement is registered in a vector
$\mathbf{x}_t = \begin{bmatrix}x_0 & x_t & y_0 & y_t & s_0 & s_t\end{bmatrix}^T$
containing initial and current $(x, y)$-coordinates and size $s$ of the marker.
Assuming a first order dynamical system $\mathbf{x}_t = \mathbf{Ax}_{t-1} + \varepsilon_t$
with linear observations $\mathbf{y}_t = \mathbf{Hx_t} + \delta_t$, we filter blob displacements
with the \emph{Kalman filter}.
Covariance matrices are taken to be proportional to the identity,
with system noise covariance~$0.01$ and observation noise~$0.1$.
Time step is taken as the inverse of the camera frame rate, i.e., $dt = 1/15$.
Definitions of matrices $\mathbf{A}$ and $\mathbf{H}$
and further detail can be found in~\cite{alymbek_thesis}.

Fig.~\ref{KalmanFilter} shows a comparison of the tracking performance
between the baseline algorithm~\cite{yamaguchi2016combining}
and our improved version in a static object holding task.
Displacement of each blob is estimated and an average displacement is shown
separately in $x$-, $y$-, and $z$-directions. The coordinate system is
placed such that the $xy$-plane is aligned with the image plane of the camera,
thus the $z$-axis is pointing from one fingertip to the other.
The baseline algorithm produces a lot of jitter due to the robot vibration
and ensuing noise in the blob detection algorithm (blue line in Fig.~\ref{KalmanFilter}).
The Kalman filter removes undesirable high-frequency jumps (orange line in Fig.~\ref{KalmanFilter}). 

\subsection{Force Tracking vs. Object Tracking}
\label{subsec:kalman_force}
\begin{figure}[t]
\vspace{.5em}
    \centering
  \includegraphics[width=\linewidth,trim={0 0 0 0},clip]{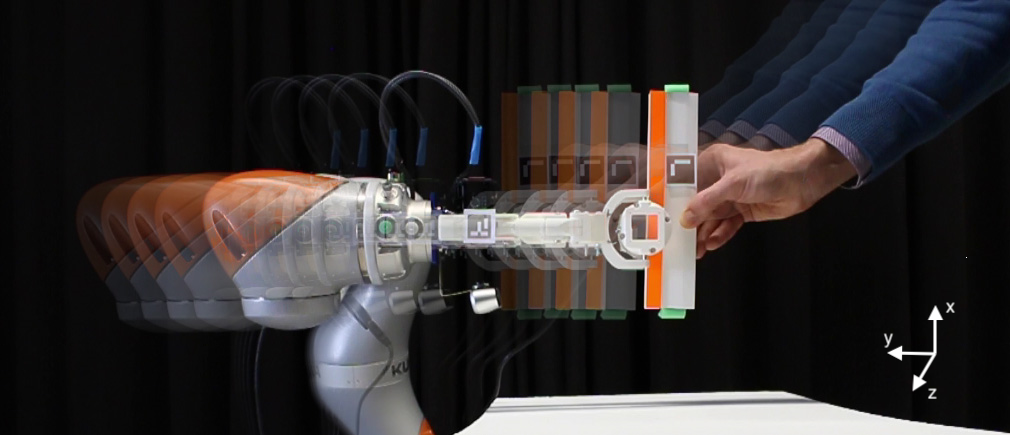}
  \caption{Robot following the motion of a human in the FollowMe task.
  ForceTrack and ObjTrack controllers compared.}
  \label{fig:pull} 
  \vspace{-2em}
\end{figure}

\begin{figure}[b]
  \vspace{-2em}
    \centering
  \subfloat[pull, force\label{1a}]{%
       \includegraphics[width=0.48\linewidth,trim={0 1em 0 0},clip]{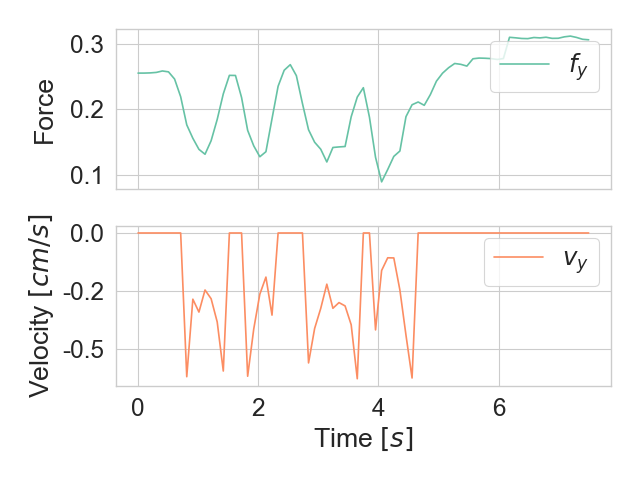}}
  \hspace{0.1em}
  \subfloat[push, force\label{1b}]{%
        \includegraphics[width=0.48\linewidth,trim={0 1em 0 0},clip]{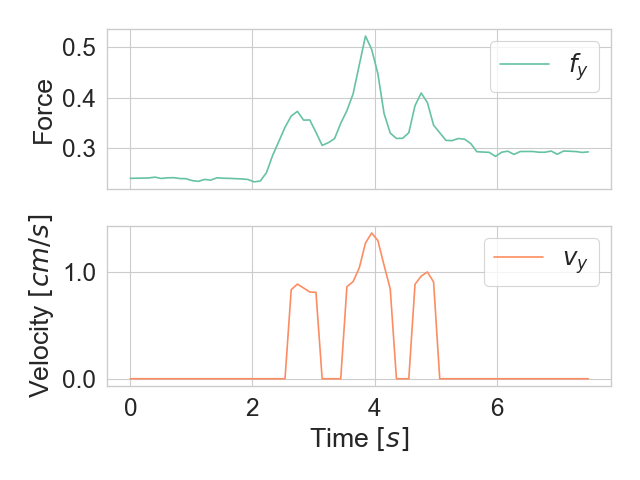}}\\
   \subfloat[pull, vision\label{1a}]{%
       \includegraphics[width=0.48\linewidth,trim={0 1em 0 0},clip]{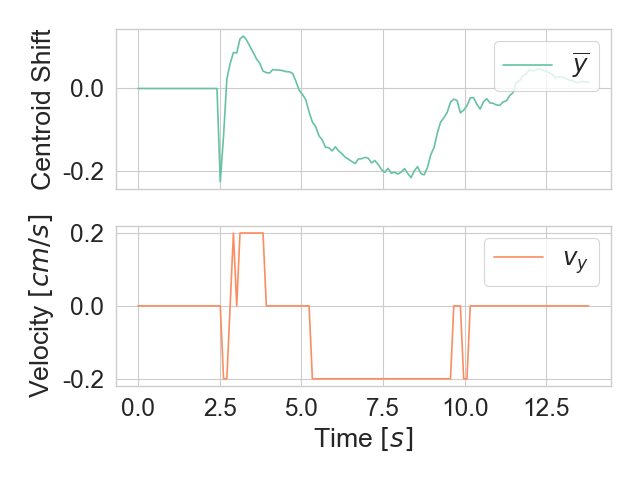}}
  \hspace{0.1em}
  \subfloat[push, vision\label{1b}]{%
        \includegraphics[width=0.48\linewidth,trim={0 1em 0 0},clip]{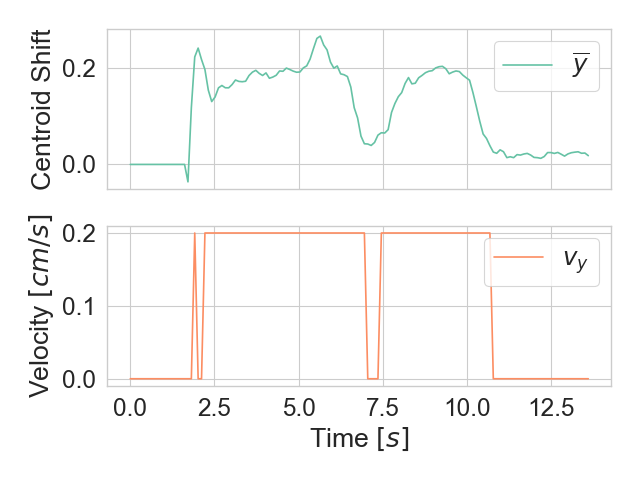}}
  \caption{ForceTrack+\textbf{SpeedCtrl} (upper row) vs. ObjectTrack (lower row) on the FollowMe task.
  	ForceTrack controller follows the input (Force) better with fewer abrupt changes.}
  \label{fig:SpeedControl} 
\end{figure}

The FingerVision sensor is unique among vision-based tactile sensors
in that it has a transparent coating and can thus utilize the raw video stream
seen through the `skin' of the sensor. But is it also useful for control?
To answer this question, we implement \emph{ForceTrack}
and \emph{ObjectTrack} controllers from~\cite{yamaguchi2017implementing}
and compare their performance on the \emph{FollowMe} task.
The human places an object inside the gripper (e.g., as shown in Fig.~\ref{fig:pull})
and then either pulls or pushes the object; the robot has to follow.
The ForceTrack skill relies on \emph{marker displacement estimation}
to move proportionally to the force applied.
The ObjectTrack skill, on the other hand, relies on \emph{optical flow estimation}
to move proportionally to the velocity of the object displacement.
Experimental results in Fig.~\ref{fig:SpeedControl}
show that the force-based approach is more robust and is easier to use.
The vision-based approach, however, results in jerkier behavior and completely fails
if a movement in the $z$-direction is required (perpendicular to the fingertip),
for such movement cannot be detected using vision.
In the following, we elaborate on the setup,
describe the details of the controller implementations, and discuss the results.

\subsubsection{ForceTrack+\textbf{SpeedCtrl}}
\label{subsec:force-based-ctrl}
This controller moves the robot end effector in the direction of the average force
measured by FingerVision.
If \mbox{$\mathbf{f}_t = \begin{bmatrix}f_t^x & f_t^y & f_t^z\end{bmatrix}^T$}
is the current force estimate, then the desired Cartesian position at the next time step is given by the rule
\mbox{$
	\mathbf{x}_{t+1} = \mathbf{x}_t + \alpha
	\cdot (\mathbf{f}_t - f_{\min})
	\cdot \mathbf{1}(\abs{\mathbf{f}_t} > \varepsilon)$}.
Parameters $\alpha, f_{\min}, \varepsilon$ are set depending on the desired responsiveness of the robot.
Notation $\mathbf{1}(b)$ stands for an indicator function that converts $b$ from True/False to $1$/$0$.
Parameter~$\alpha$ controls the movement speed, which we call \textbf{SpeedCtrl} feature.
We set $\alpha = (v_{\max} - v_{\min}) / (f_{\max} - f_{\min})$
to control the range of commands $\mathbf{x}_{t+1}$ through the velocity range.

\subsubsection{ObjectTrack}
This controller tries to keep the object in the center of the field of view of the FingerVision camera.
Assume the object is moved in the $y$-direction (see Fig.~\ref{fig:pull}).
If $M_t$ is the current object area (computed using image moments~\cite{pawlak2006image}
as described in~\cite{yamaguchi2017implementing}),
and~$(\bar{x}_t, \bar{y}_t)$ are the coordinates of the object center in the image plane,
then the desired Cartesian position is set as
\begin{equation} 
\label{eq:following-prox} 
\begin{aligned} 
 x_{t+1} &= x_t + \min{\{\overline{x}, x_{\max}\}} \cdot \mathbf{1}(\abs{\overline{x}} > \overline{x}_{\epsilon}),\\ 
 y_{t+1} &= y_t + \Delta \left(\mathbf{1}(M_t \leq M_{\epsilon}) - \mathbf{1}(M_t > M_{\epsilon}) \right)\\ 
\end{aligned} 
\end{equation}
where $\Delta$ is the sensitivity to the displacement of the object inside the gripper.
Parameters $x_{\max}$ and $y_{\max}$ are the maximum allowed deviations
in the $x$ and $y$ directions respectively,
and thresholds $\overline{x}_{\epsilon}$, $\overline{y}_{\epsilon}$, $M_{\epsilon}$
prevent false positive controller activations.
For moving in the $x$-direction, equations in~\eqref{eq:following-prox} should be swapped.

\subsubsection{ForceTrack+\textbf{SpeedCtrl} vs ObjectTrack}
Fig.~\ref{fig:SpeedControl} compares ForceTrack (upper) and ObjectTrack(lower) controllers
under two conditions: pulling (left) and pushing (right).
Each plot shows the input signal at the top (either force or centroid shift)
and the command generated from it at the bottom (end-effector velocity in both cases).
Note that the force is not measured in Newtons but is proportional to the markers displacements,
as in~\cite{yamaguchi2016combining}.
ForceTrack controller \emph{with} SpeedCtrl was consistently better than \emph{without}.
As seen from Fig.~\ref{fig:SpeedControl}, the force-based controller follows the input signal more faithfully.
Additionally, it is more robust with respect to the object used~\cite{alymbek_thesis},
in contrast to the vision-based controller.
Based on our experiments, \emph{the force-based controller should be preferred for contact-tracking tasks},
such as the FollowMe task showcased here.

\subsection{Torque-Driven Arm Rotation \textbf{ArmRot}}
\label{subsec:trq-arm-rot}
\begin{figure}[t]
\vspace{.25em}
    \centering
  \includegraphics[width=1.0\linewidth,trim={0 1em 0 1em},clip]{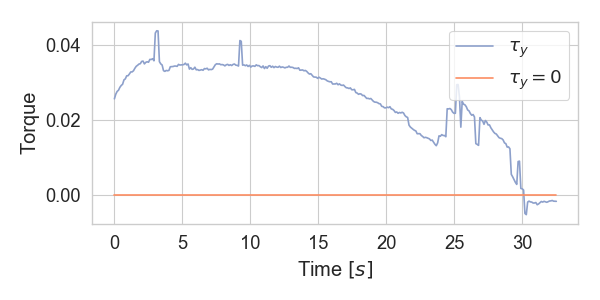}
  \caption{\textbf{ArmRot} skill. An asymmetric object held in-fingers produces torque (blue).
  The robot rotates the arm to reach zero torque (orange).}
  \label{fig:ArmRotation} 
  \vspace{-1.5em}
\end{figure}
A torque estimation procedure for FingerVision was proposed in~\cite{yamaguchi2017implementing}.
However, it was not used for control.
Here, we demonstrate a controller that can successfully utilize such torque information.
Fig.~\ref{fig:ArmRotation} demonstrates a potential use case: an asymmetric stick with a heavy head
is grasped by the tail and held horizontally in the beginning. The torque created by the stick in the fingers
is detected by FingerVision and is used in a feedback loop to drive sensor readings to zero.

\subsection{Handover with Leaky Integrator \textbf{LeakyInt}}
\label{subsec:leaky}
\begin{figure}[t]
\vspace{.25em}
    \centering
  \includegraphics[width=1.0\linewidth,trim={0 1em 0 1em},clip]{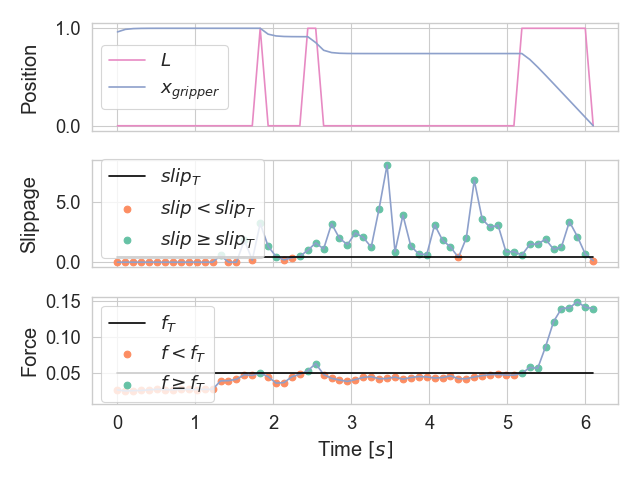}
  \caption{\textbf{LeakyInt} skill triggers gripper closure when both force and slippage signals are active.
  In general, any external signal can be used to trigger the leaky integrator.}
  \label{fig:LeakyIntegrator} 
    \vspace{-1.5em}
\end{figure}
Gentle opening and closing of the gripper is the most basic skill one can expect.
We implement the leaky integrator as described in~\cite{veiga2018hand} for the FingerVision.
The idea is to follow an external input with some inertia. More concretely,
the position command $x_t$ for the gripper is computed as $x_t = \alpha x_{t-1} - (1 - \alpha)L$
where $\alpha$ controls how fast the gripper reacts and $L$ determines the set point.
Note that $L$ can be any external input. For example, Fig.~\ref{fig:LeakyIntegrator} demonstrates
how a combination of slippage and force can be used to trigger  $L$ (pink line in the top subplot).
Since $L$ is only activated when both Slippage (middle subplot, green points mean `active',
red points mean `not active') and Force (bottom subplot, the same color convention) are active,
the gripper is closing when an object is detected between the fingers
and at the same time the object is touching the sensor (pink line).

\subsection{In-Hand Object Rotation \textbf{InHandRot}}
\label{subsec:in-hand-rot}
\begin{figure}[t]
\vspace{.5em}
    \centering
  \includegraphics[width=1.0\linewidth,trim={0 1em 0 1em},clip]{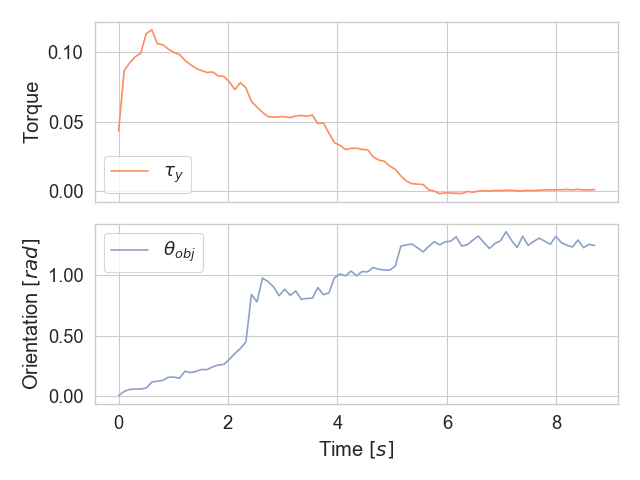}
  \caption{\textbf{InHandRot}. A pen is left to rotate under its own weight between the fingers from the position
  parallel to the ground. Although zero torque was detected (upper plot), the final pen orientation was
  less than $90$ degrees which means the pen got stuck.
  This is a typical problem of torque estimation for light objects with FingerVision.}
  \label{fig:InHandRotation}
  \vspace{-1.5em}
\end{figure}
We compare two approaches to rotating an object inside a parallel gripper:
based on torque estimation and based on slip detection.
Both approaches perform well, the choice may depend on the application.
The slip-detection approach was better for letting objects rotate under their own weight
because torque estimation was unreliable for light objects and small torques.
Fig.~\ref{fig:InHandRotation} shows the torque signal used for rotating a pen inside the gripper.
We slowly open the gripper using the leaky integrator till the value of torque goes to zero.

\section{LEARNING-BASED SKILLS}
\label{sec:learning-based-skills}
In the previous section, improvements over baseline controllers
and a few novel analytic controllers utilizing the Markers modality
were described (see Table~\ref{table:tactile_skills}).
This section is aimed to demonstrate the feasibility of utilizing the raw input
in the form of marker displacement data and optical flow estimation data
for controlling the robot.
To that end, two example applications are presented:
\textbf{\textit{ForceLearn}} which puts emphasis on force estimation data
and \textbf{\textit{StirLearn}} which puts emphasis on the optical flow estimation data.

\subsection{\textbf{\textit{ForceLearn}}: Learning to Press with a Given Force}
\label{subsec:force-learn}
\begin{figure}[b]
\vspace{-2em}
    \centering
  \subfloat[Regression\label{1b}]{%
        \includegraphics[width=0.48\linewidth]{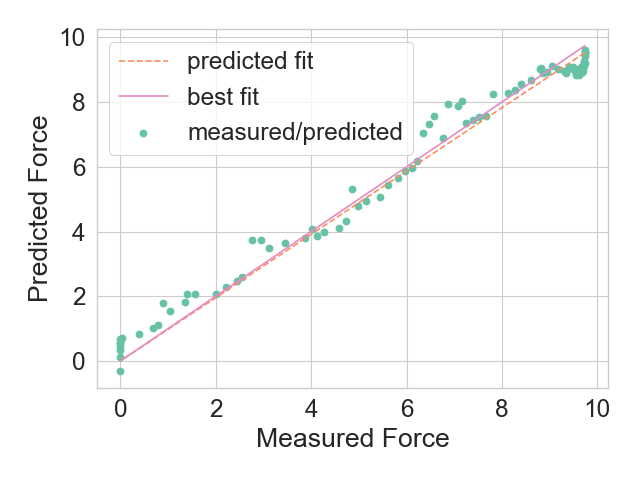}}
  \hspace{0.1em}
  \subfloat[Prediction\label{1c}]{%
        \includegraphics[width=0.48\linewidth]{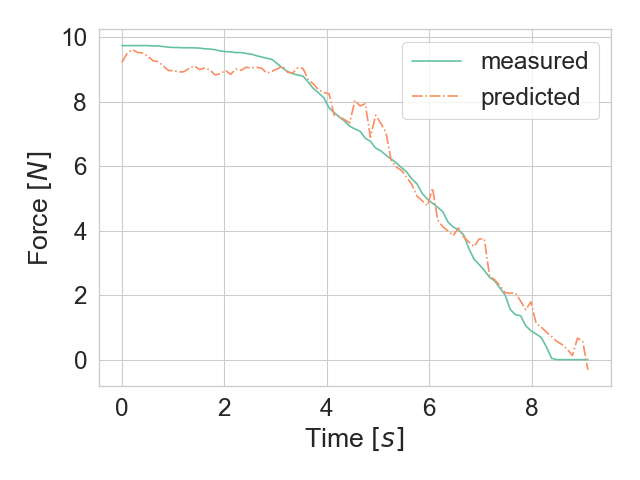}}
  \caption{\textbf{\textit{ForceLearn}}. Training (left) and testing (right) of a learned predictor
  of the force from marker displacements. Prediction is accurate and can be used in downstream tasks.}
  \label{fig:ForceLearning} 
\end{figure}
Associating marker displacements with the exerted force is a non-trivial task.
On one hand, marker displacements should be proportional to the force, at least for small deformations.
On the other hand, the displacements depend on other factors, such as the material of the objects in contact and
the kinematic configuration of the problem.
Learning-based approaches may potentially be sufficiently powerful to extract invariant information
which can be used for control. We provide a proof-of-concept demonstration that this is indeed the case
and machine learning can be effective for encoding the mapping from sensor readings to force values.

We train a regression model to predict force from marker displacements.
Our setup consists of an electronic scale that delivers data over ROS
and the robot pushing a stick held in-between the fingers against the scale.
Fig.~\ref{fig:ForceLearning} shows the results obtained
using kernel ridge regression with radial basis functions.
The left figure shows measured vs. predicted values;
the right figure shows the output of the predictor obtained in a real-time test while releasing a button.
Using $20$ pressing movements each of duration $5$ sec recorded at $15$ Hz
was sufficient to predict force values with the measurement accuracy.
However, one has to stress that the learned mapping is object-dependent
due to the use of silicone as the skin in FingerVision: different materials behave differently
when in contact with silicone. For example, the FingerVision gets stuck on glass.
It is not yet clear what a general object-agnostic tactile sensor should look like.

\subsection{\textbf{\textit{StirLearn}}: Density and Texture through Stirring}
\label{subsec:stir-learn}
\begin{table}[t]
\vspace{.5em}
\caption{\textbf{\textit{StirLearn}}: Classifying substances by stirring.}
\label{table:density-results} 
\centering 
\begin{tabular}{|c|c|c|c|c|}
\hline 
          & precision & recall & f1-score & support \\ \hline 
flour & 1.00 & 0.94 & 0.97 & 16 \\ \hline 
sugar & 1.00 & 1.00 & 1.00 & 16 \\ \hline 
peas & 0.94 & 1.00 & 0.97 & 16 \\ \hline 
avg/total & 0.98 & 0.98 & 0.98 & 48 \\ \hline 
\end{tabular} 
\vspace{-2em}
\end{table} 
As the final demonstration, we showcase the use of all input modalities in a challenging prediction task.
We consider a problem of discerning substances that have different density and texture,
such as flour, sugar, and peas, based on tactile interaction with them.
This problem setup is inspired by~\cite{elbrechter2015discriminating},
where a system was trained to discriminate liquids
of varying viscosity by detecting surface changes with a depth camera.
In our case, the features are not based on visual observations but rather on tactile sensations.
As the feature vector, we use all available information, i.e.,
deviation of each marker, centroid $\left\{ \overline{x}, \overline{y} \right\}$,
object orientation~$\theta$ and area~$M$.

A data set consisting of $120$ trials was collected via stirring substances with a wooden stick
using a set of $8$~pre-defined stirring movements and grasping settings.
A training set and a test set were created, comprised of $72$ and $48$~trials, respectively.
A multi-layer perceptron (MLP) with $3$~hidden layers, $10$ neurons per layer, and logistic activations
was used.
The evaluation metrics of the trained MLP are provided in \Cref{table:density-results}.
A virtually ideal classifier could be obtained.

\section{HUMAN-ROBOT COLLABORATION IN ARCHITECTURAL ASSEMBLY TASKS}
\label{sec:assembly}
\begin{figure*}[t]
	\centering
	\includegraphics[width=.94\textwidth]{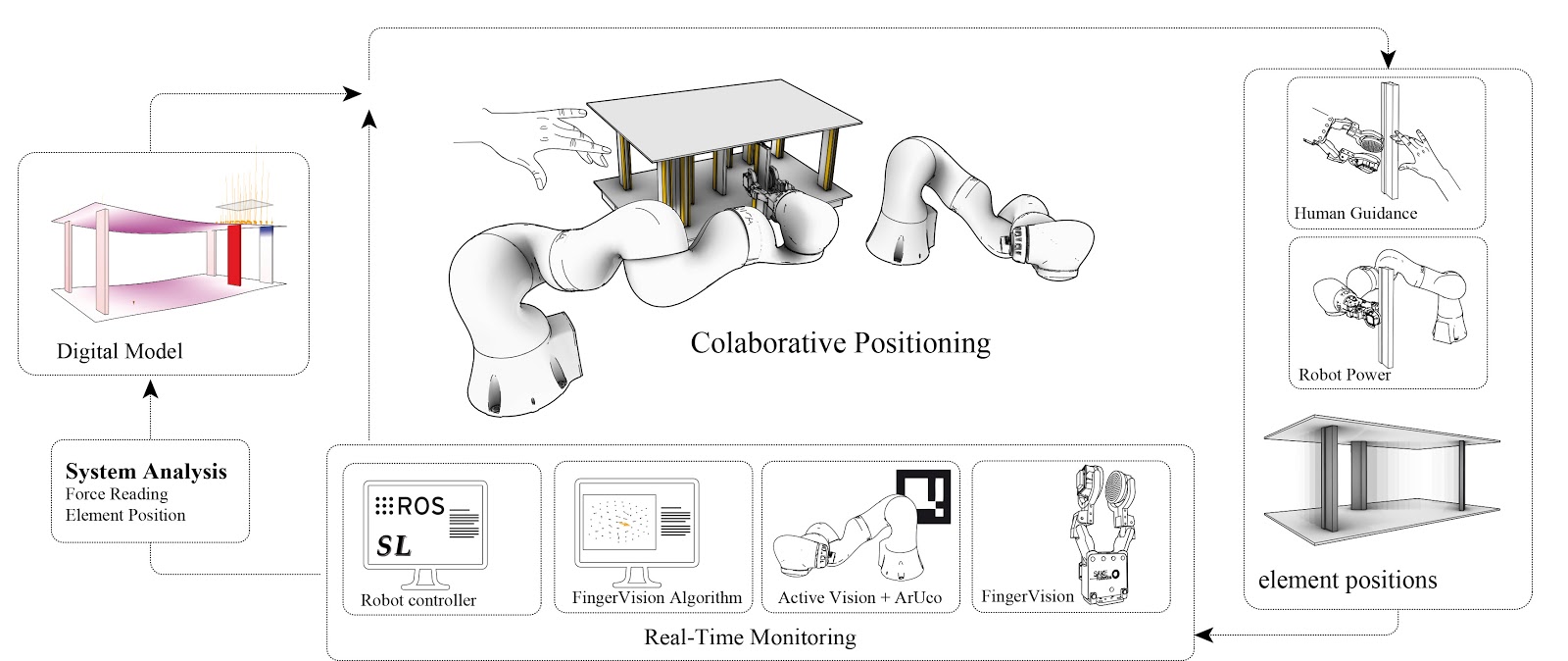}
	\vspace*{-4pt}
	\caption{A prototype pipeline for collaborative architectural assembly.
		The outer feedback loop shows the data flow from the design software
		to the robot and from the sensors registering the actual configuration back to the design model.
		The assembly process detailed in Fig.~\ref{fig:AssemblyTask} and comprised of
		skills from Table~\ref{table:tactile_skills} allows for interaction via
		the Handover+\textbf{LeakyInt} (Sec.~\ref{subsec:leaky})
		and ForceTrack+\textbf{SpeedCtrl} (Sec.~\ref{subsec:force-based-ctrl}) skills.
		See Fig.~\ref{fig:SpeedControl} and the accompanying video for details.}
	\vspace*{-10pt}
	\label{fig:assembly}
\end{figure*}
Previous two sections introduced a variety of skills based on tactile feedback.
In this section, we want to demonstrate the utility
of developing such a repertoire of skills by showing how they can be combined together
to solve a complex contact-rich task.
As an application, we consider \emph{collaborative architectural assembly}
based on digital design models~\cite{wibranek_caad19}.
The common practice today requires an architect to precisely define in advance
both the local and global geometry of a structure to be built.
However, unforeseen changes often occur after the construction has been started.
Especially interesting are the cases where the changes are not due to robot mistakes but rather
indicate architect's changing design goals in an interactive fashion,
such as in the \emph{collaborative positioning} task showcased below.

Fig.~\ref{fig:assembly} shows the model used in our experiments.
A digital model describing a desired structure is passed to the robot for assembly.
Locations of the vertical load-bearing elements are not pre-programmed
but rather determined by the robot through tactile sensing online as the locations of the highest load.
Subsequently, the observed state of the erected structure is passed
back to the design software to update the plan. Interaction with the robot is enabled through
the tactile feedback and can be used to guide the robot and reposition the elements.
See the accompanying video for details.

External lighting and bright-colored objects were used in the assembly task.
Experiments exposed high sensitivity of the vision-based
blob detection algorithms~~\cite{yamaguchi2016combining}
to lighting conditions and object color.
Installing a light source at the wrist could potentially solve these problems.

\section{CONCLUSION}
Tactile-enabled applications have been a long-standing vision in robotics~\cite{cutkosky2008force}.
In the recent years, the combination of inexpensive hardware with advances in machine learning
are providing a unique opportunity to experiment with novel designs and applications of tactile sensors.
In this paper, we attempted to build a library of useful robot behaviors
by utilizing tactile feedback (see Sec.~\ref{sec:library}).
To accomplish that, we created a modified version of the FingerVision sensor~\cite{yamaguchi2016combining}
that suits our robot hardware and provides a few design enhancements (see Sec.~\ref{sec:building}).
We further extended the existing software framework around the FingerVision~\cite{yamaguchi2017implementing}
with a suite of improved and novel tactile skills (see Sec.~\ref{sec:improved-and-novel}).
Beyond hand-designed controllers, we for the first time demonstrated the feasibility of using
the raw sensory data from the Finger vision to learn skills
such as pressing with a specified amount of force (Sec.~\ref{subsec:force-learn})
and identifying substances through stirring (Sec.~\ref{subsec:stir-learn}).
Finally, we showcased a potential future application of tactile sensing
in interactive architectural assembly~\cite{wibranek_caad19}
based on human-robot collaboration.

\addtolength{\textheight}{-12cm}   

%

\section*{ACKNOWLEDGMENT}

We thank Christian Betschinske for his great help in building our FingerVision sensor
and Olivier Stoos for creating the architectural assembly model.
A.S. thanks Lufthansa Industry Solutions AS GmbH for covering the travel costs.

\bibliographystyle{ieeetran}
\bibliography{ieeeabrv,references}
\end{document}